\documentclass[letterpaper]{article} 
\usepackage{aaai2026}  
\usepackage{times}  
\usepackage{helvet}  
\usepackage{courier}  
\usepackage[hyphens]{url}  
\usepackage{graphicx} 
\usepackage{natbib}  
\usepackage{caption} 
\usepackage{amsmath}
\usepackage{amssymb}
\usepackage{algorithm}
\usepackage{algorithmic}
\usepackage[table]{xcolor}         
\usepackage[dvipsnames]{xcolor}
\usepackage{bm}
\usepackage{xspace}
\usepackage{booktabs}
\usepackage{colortbl}
\usepackage{marvosym}
\definecolor{firstcolor}{HTML}{BDE6CD}
\definecolor{secondcolor}{HTML}{E2EEBC}
\definecolor{thirdcolor}{HTML}{FFF8C5}

\usepackage{newfloat}
\usepackage{listings}
\DeclareCaptionStyle{ruled}{labelfont=normalfont,labelsep=colon,strut=off} 
\floatstyle{ruled}
\newfloat{listing}{tb}{lst}{}
\floatname{listing}{Listing}
\title{FoundationSLAM: Unleashing the Power of Depth Foundation Models for \\ End-to-End Dense Visual SLAM}
\author {
    Yuchen Wu\textsuperscript{\rm 1},
    Jiahe Li\textsuperscript{\rm 1},
    Fabio Tosi\textsuperscript{\rm 2},
    Matteo Poggi\textsuperscript{\rm 2},
    Jin Zheng\textsuperscript{\rm 1,3},
    Xiao Bai\textsuperscript{\rm 1}\footnote{Corresponding author: Xiao Bai (baixiao@buaa.edu.cn)}
}
\affiliations {
    \textsuperscript{\rm 1}School of Computer Science and Engineering, State Key Laboratory of Complex Critical
 Software Environment, Jiangxi Research Institute, Beihang University\\
    \textsuperscript{\rm 2}University of Bologna \\
    \textsuperscript{\rm 3}State Key Laboratory of Virtual Reality Technology and Systems, Beijing, China\\
    wuyuchen@buaa.edu.cn, lijiahe@buaa.edu.cn, fabio.tosi5@unibo.it
}

\usepackage{bibentry}

\begin{document}

\maketitle

\begin{abstract}
We present FoundationSLAM, a learning-based monocular dense SLAM system that addresses the absence of geometric consistency in previous flow-based approaches for accurate and robust tracking and mapping.
Our core idea is to bridge flow estimation with geometric reasoning by leveraging the guidance from foundation depth models. 
To this end, we first develop a Hybrid Flow Network that produces geometry-aware correspondences, enabling consistent depth and pose inference across diverse keyframes. 
To enforce global consistency, we propose a Bi-Consistent Bundle Adjustment Layer that jointly optimizes keyframe pose and depth under multi-view constraints. Furthermore, we introduce a Reliability-Aware Refinement mechanism that dynamically adapts the flow update process by distinguishing between reliable and uncertain regions, forming a closed feedback loop between matching and optimization.
Extensive experiments demonstrate that FoundationSLAM achieves superior trajectory accuracy and dense reconstruction quality across multiple challenging datasets, while running in real-time at 18 FPS, demonstrating strong generalization to various scenarios and practical applicability of our method.
\end{abstract}


\section{Introduction}
\label{sec:intro}
Simultaneous Localization and Mapping (SLAM) is a fundamental problem in computer vision and robotics, enabling autonomous agents to perceive and navigate unknown environments. Recent advances in learning-based SLAM systems have demonstrated impressive results by leveraging dense optical flow as a unified representation for tracking and mapping. Among them, DROID-SLAM~\cite{teed2021droid} and a series of its extended variants~\cite{zhang2023go, mod-slam, HI-SLAM} have established new performance baselines in multiple benchmarks, showcasing the potential of flow-based dense SLAM approaches in theory and practice.

\begin{figure}[t]
\centering
\includegraphics[width=\linewidth]{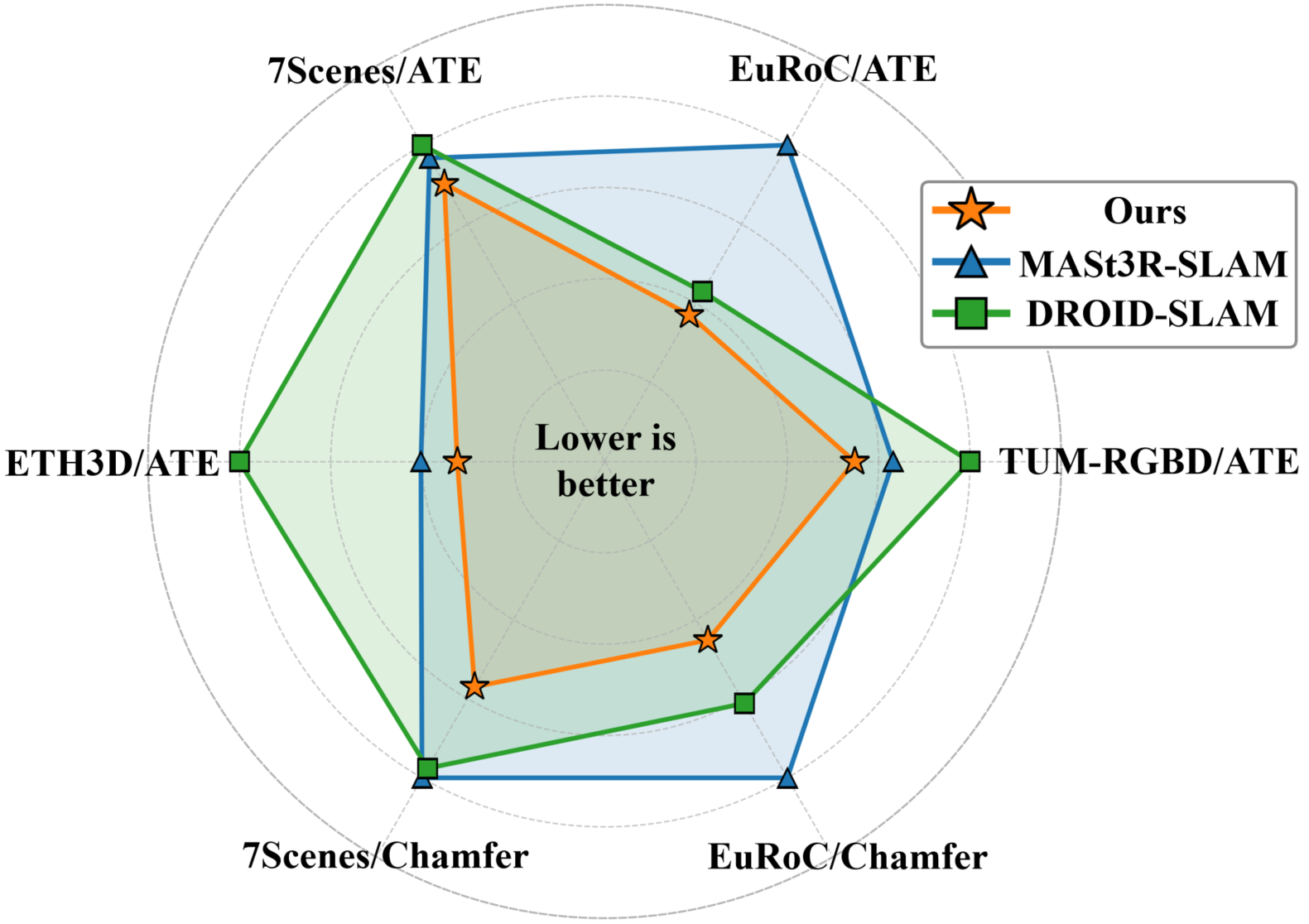}
\caption{SLAM Performance Comparison. Radar plot shows normalized ATE (TUM, EuRoC, 7Scenes, ETH3D) and Chamfer distance (7Scenes, EuRoC). Our method (orange star) achieves optimal performance on any metrics.}
\label{fig:radar}
\end{figure}

Despite recent progress, existing flow-based monocular dense SLAM systems still perceive and reconstruct scenes solely through pixel-wise correspondences from the 2D optical flow, resulting in a lack of geometric consistency in both tracking and mapping. As a result, the reconstructed depth may exhibit structural artifacts, layered ambiguities, or incomplete geometry, ultimately degrading pose accuracy and reconstruction quality.

Specifically, two key factors are primarily behind this limitation. First, dense correspondence estimation is performed solely in image space and lacks awareness of underlying scene geometry, leading to structurally inconsistent matches across views, especially in textureless and ambiguous regions. Second, current systems lack explicit enforcement of multi-view geometric constraints during optimization and dedicated mechanisms to refine flow predictions based on these constraints, resulting in accumulated errors during optimization. These issues call for a tightly coupled framework where geometric priors direct the correspondence estimation, and multi-view optimization in turn guides refinement.

To this end, we propose FoundationSLAM, a monocular dense SLAM framework that integrates geometric guidance with multi-view constrained optimization into a fully differentiable pipeline. 
Our approach leverages the strong priors encoded in foundation depth models to guide flow matching under challenging conditions, and introduces a novel Bi-Consistent Bundle Adjustment Layer that jointly refines depth and pose while enforcing multi-view consistency. Furthermore, a residual-based refinement mechanism enables the system to identify unreliable predictions and use geometric priors for targeted correction, thereby achieving robust and interactive SLAM across diverse scenarios.

We demonstrate the effectiveness of FoundationSLAM across multiple challenging benchmarks, including TUM-RGBD~\cite{sturm2012benchmark}, EuRoC~\cite{burri2016euroc}, 7Scenes~\cite{glocker2013real}, and ETH3D~\cite{schops2019bad}, outperforming existing approaches, as shown in Figure \ref{fig:radar}. Our main contributions are:
\begin{itemize}
  \item We propose a Hybrid Flow Network that leverages geometric priors from foundation models for robust correspondence estimation. A Reliability-Aware Refinement module further improves predictions by selectively correcting unreliable flow based on optimization residuals.

  \item We propose a Bi-Consistent Bundle Adjustment Layer that jointly optimizes dense depth and pose with multi-view residuals. By enforcing bidirectional consistency, our formulation improves tracking and reconstruction accuracy in challenging scenes.
  
  \item Extensive experiments on standard SLAM benchmarks demonstrate that our method outperforms previous monocular dense SLAM systems in both trajectory accuracy and reconstruction quality, while running in real time at 18 FPS.
\end{itemize}

\section{Related Work}

\textbf{Matching-based SLAM.}
Classical and modern SLAM systems often adopt a keyframe-based matching paradigm using inter-frame correspondences to jointly optimize camera poses and scene geometry. Early systems (e.g., ORB-SLAM~\cite{mur2017orb}, DSO~\cite{dso}) rely on handcrafted pipelines involving tracking, mapping, and bundle adjustment. With the rise of deep learning, end-to-end SLAM methods such as DROID-SLAM~\cite{teed2021droid} and DPVO~\cite{teed2023deep} incorporate learnable feature extractors and differentiable optimization modules for dense, robust tracking.
A key advantage of these methods lies in the tight coupling between front-end perception and back-end optimization, where matching predictions and optimization mutually inform each other, leading to accurate, consistent SLAM even in challenging scenarios. However, these methods still estimate pixel-wise correspondences solely based on local correlation, without incorporating explicit geometric priors. This makes them prone to matching failures in textureless or ambiguous regions. More critically, the estimated flow is not constrained to be consistent across views, leading to structural inconsistencies and degraded reconstruction quality over long sequences.

\begin{figure*}[!t]
\centering
\includegraphics[width=\linewidth]{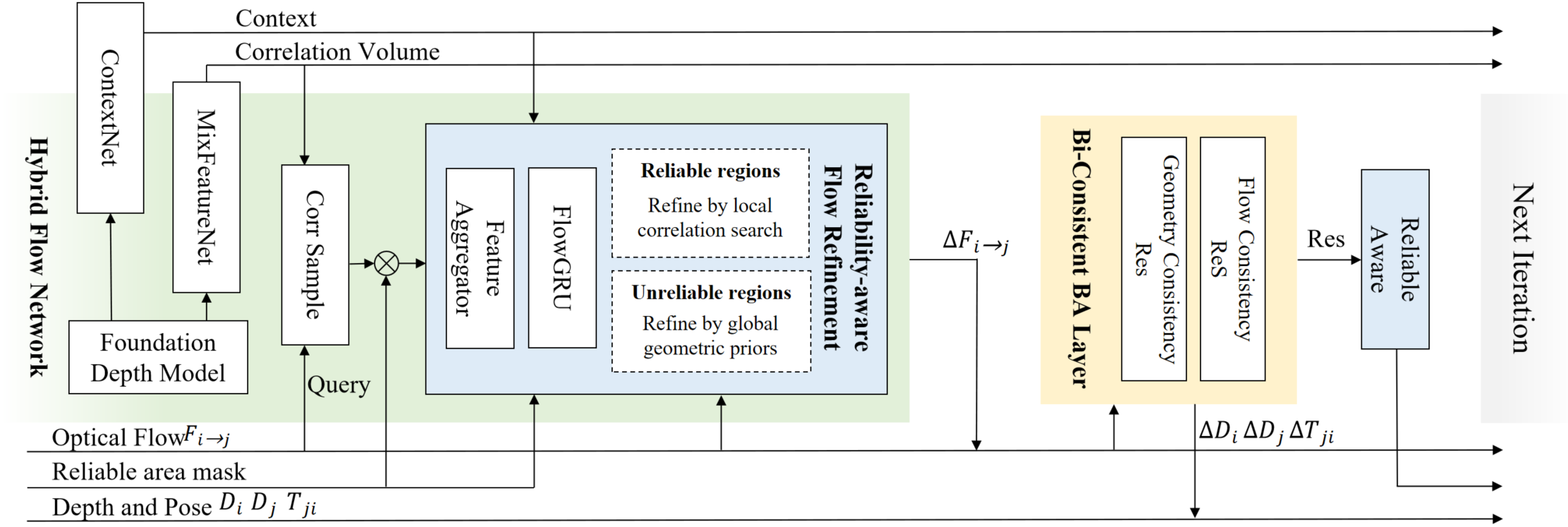}
\caption{Method Overview. Given a pair of keyframes, we estimate dense optical flow using a hybrid network fusing geometry-aware features from a foundation depth model. Predicted flow is iteratively refined via Flow GRU, guided by context features and learned reliability masks. Refined flow drives a Bi-Consistent BA Layer jointly optimizing keyframe depth and pose using flow and geometry consistency residuals. Optimization feedback updates flow reliability in closed-loop manner. This process unrolls over multiple iterations to progressively improve accuracy and consistency.}
\label{fig:framework}
\end{figure*}

\noindent\textbf{Hybrid SLAM with Global Scene Representations.} To improve global consistency, several recent works incorporate explicit 3D representations such as NeRF~\cite{imap, nice-slam, johari2023eslam} or 3D Gaussian Splatting~\cite{splatam, yan2024gs, matsuki2024gaussian} into SLAM pipelines~\cite{tosi2024nerfs}. These methods use a matching-based frontend for pose tracking and simultaneously optimize a global scene representation. This hybrid structure enables dense and globally consistent reconstructions and even supports loop closure.
Despite their benefits, these systems suffer from key drawbacks: they often require scene-specific optimization, assume known intrinsics, and demand high computational cost. More importantly, the global representation is updated independently of the pose tracker, weakening the feedback between perception and optimization. Compared to tightly coupled systems, this loose interaction further reduces the ability to correct or guide front-end predictions using multi-view consistency.

\noindent\textbf{SLAM from 3D Reconstruction Models.} Foundation models for 3D reconstruction, such as DUSt3R~\cite{wang2024dust3r} and MASt3R~\cite{leroy2024grounding}, have shown remarkable capabilities in predicting dense scene geometry and relative poses from as few as two uncalibrated images. These priors are increasingly used in SLAM pipelines such as MASt3R-SLAM~\cite{Murai_2025_CVPR}, VGGT-SLAM~\cite{maggio2025vggt_slam}, and VGGT~\cite{wang2025vggt}, where per-frame geometry predictions provide additional guidance for inter-frame matching. However, these systems typically predict geometry on a per-frame or per-pair basis, and the frontend priors are not explicitly refined or guided by the backend optimization.
As a result, these systems often suffer from inaccurate priors in challenging regions and lack a mechanism to correct them through joint optimization. Other approaches (e.g., SLAM3R~\cite{liu2025slam3r}) discard back-end optimization entirely and directly fuse point clouds from foundation models, achieving efficiency but at the cost of robustness and long-term accuracy.

This work aims to unify these strengths by embedding geometry-aware priors into a tightly integrated optimization framework that enforces multi-view geometric consistency while retaining efficiency and generalization.

\section{Method}

FoundationSLAM addresses the lack of geometric consistency in flow-based SLAM systems through a fully differentiable and tightly coupled framework. As shown in Figure~\ref{fig:framework}, our method consists of three key components: (1) a Hybrid Flow Network that injects structural awareness into correspondence estimation via geometric priors from foundation depth models, (2) a Bi-Consistent Bundle Adjustment Layer that jointly refines depth and pose while enforcing multi-view consistency, and (3) a Reliability-Aware Refinement mechanism that leverages optimization residuals to guide flow correction. Together, these components form a closed-loop system to achieve robust and consistent monocular dense SLAM.

\subsection{Hybrid Flow Estimation with Geometry Prior}

A core limitation of existing flow-based SLAM systems is that dense correspondences are estimated independently for each image pair, without enforcing consistency across views. This lack of structural awareness often leads to geometrically inconsistent matches, especially in low-texture or ambiguous regions, ultimately degrading the accuracy of downstream optimization. To address this, we propose a Hybrid Flow Network that injects geometric priors from foundation depth models into the correspondence estimation process. These priors encode global scene structure and enable the flow network to produce more consistent and reliable matches across multiple viewpoints.

\noindent\textbf{Backbone Design.} Inspired by FoundationStereo's~\cite {foundationstereo} effective integration of depth priors~\cite{depthanythingv2}, we design a dual-branch architecture: (1) a \textit{Geometric Prior Branch} utilizing the frozen FeatureNet encoder from FoundationStereo to provide stable geometric features learned from diverse real-world imagery, and (2) a \textit{Task-Specific Adaptation Branch} with trainable CNN layers mirroring parts of FeatureNet, optimized for monocular SLAM data association challenges. Features from the two branches are fused via $3{\times}3$ convolution followed by residual layers, yielding the final matching descriptor.
We additionally incorporate the frozen ContextNet from FoundationStereo for context with rich geometric priors. These pretrained modules are used purely as convenient sources of geometry-guided features and remain fixed during training.

\noindent\textbf{Flow Estimation Process.} Operating over a dynamically maintained keyframe graph, MixFeatureNet extracts fused matching features for target frames and neighbors, while ContextNet provides context feature. A Feature Aggregator processes these inputs with initial flow estimates, feeding into a Flow GRU module that iteratively predicts flow updates ($\Delta\emph{F}$) and confidence maps ($\omega$). The integration with optimization feedback is detailed below.

\subsection{Bi-Consistent Bundle Adjustment Layer}

While modern SLAM systems combine flow prediction and optimization in a unified pipeline~\cite{teed2021droid, teed2023deep}, flow is typically estimated independently for each frame pair, without enforcing consistency across multiple views. This frame-wise decoupling limits the ability to recover globally coherent structure and leads to geometric inconsistencies.

To address this, we propose the Bi-Consistent Bundle Adjustment (BA) Layer, which introduces explicit multi-view geometric supervision into the optimization loop. By incorporating both flow alignment and geometric consistency residuals, it strengthens the connection between correspondence prediction and scene-level optimization, improving robustness and global coherence.

\noindent\textbf{Flow Consistency Residual.}
Given a pixel $\mathbf{u}_i$ in frame $i$, a 3D point is reconstructed using the estimated depth $D_i$, transformed to frame $j$ using relative pose $T_{ji}$, and projected to the image plane:
\begin{equation}
\mathbf{u}_{\text{proj}} = \pi(T_{ji} \cdot \pi^{-1}(\mathbf{u}_i, D_i)).
\end{equation}

We minimize the residual to the predicted correspondence:
\begin{equation}
\mathcal{L}_{\text{flow}} = \| \mathbf{u}_{\text{proj}} - (\mathbf{u}_i + \emph{F}_{i \rightarrow j}) \|_1.
\end{equation}
This residual supervises the alignment between depth and flow, but only reflects consistency from frame $i$'s perspective, which still lacks the multi-view geometry consistency.

\noindent\textbf{Geometry Consistency Residual.}
To enforce explicit multi-view consistency, we introduce a symmetric constraint: for each $\mathbf{u}_i$ in frame $i$, we project its reconstructed 3D point to frame $j$, and then sample the depth $D_j$ at the projected location $\mathbf{u}_j$. Then, we check whether it supports back-projection to the original pixel to get the residual:
\begin{align}
\mathbf{u}_j &= \pi(T_{ji} \cdot \pi^{-1}(\mathbf{u}_i, D_i)), \\
\mathbf{u}_i^{\text{back}} &= \pi(T_{ij} \cdot \pi^{-1}(\mathbf{u}_j, D_j)), \\
\mathcal{L}_{\text{geo}} &= \| \mathbf{u}_i^{\text{back}} - \mathbf{u}_i \|.
\end{align}
This residual penalizes geometric misalignment when frame $j$ does not geometrically support the depth prediction from frame $i$. In practice, we compute $\mathcal{L}_{\text{geo}}$ only for pixels with $\mathcal{L}_{\text{geo}} < \tau$ (with $\tau = 1$ pixel) to avoid enforcing agreement across occlusions or depth discontinuities.

To balance the influence of the two residuals, we introduce a confidence map $\omega(\mathbf{u}_i)$ that reflects the reliability of local flow predictions. The final loss combines both terms:

\begin{equation}
\mathcal{L}_{\text{BA}} = \sum_{\mathbf{u}_i \in \Omega} \left( \omega(\mathbf{u}_i) \cdot \mathcal{L}_{\text{flow}}(\mathbf{u}_i) + \left(1 - \omega(\mathbf{u}_i)\right) \cdot \mathcal{L}_{\text{geo}}(\mathbf{u}_i) \right)
\end{equation}
where $\Omega$ is the set of valid pixels satisfying the consistency threshold. We minimize $\mathcal{L}_{\text{BA}}$ using Gauss-Newton optimization. Each iteration performs once flow update followed by twice BA, we compute Jacobians with respect to both depth and pose and solve for incremental updates $\Delta D$ and $\Delta T$. This bidirectional formulation integrates local matching cues with multi-view geometric constraints, leading to more consistent and robust SLAM optimization, especially in scenes with wide baselines, occlusion, or low texture.

\subsection{Reliability-aware Flow Refinement}
\label{sec:reliability}
Accurate optical flow estimation is critical for robust SLAM, yet it often degrades in regions affected by occlusion, low texture, repetitive patterns, or wide-baseline motion. When uncorrected, these unreliable flows introduce errors into depth and pose estimates, compromising the global consistency of the reconstruction.

To mitigate this, we propose Reliability-aware Flow Refinement, a mechanism that dynamically adapts the refinement behavior based on geometric residuals from the Bi-Consistent BA Layer. Specifically, we construct a pixel-wise reliability mask that guides the flow update process by identifying regions of high and low confidence.

\noindent\textbf{Edge-wise Flow Reliability.}
Given a co-visible keyframe pair $(I_i, I_j)$, we first compute a local reliability indicator $M^{\text{edge}}_{i \rightarrow j}(\mathbf{u}) \in \{0,1\}$ based on forward projection residual:
\begin{equation}
M^{\text{edge}}_{i \rightarrow j}(\mathbf{u}) =
\begin{cases}
1 & \text{if } \mathcal{L}_{\text{proj}}(\mathbf{u}) < \tau_{\text{edge}}, \\
0 & \text{otherwise},
\end{cases}
\end{equation}
where $\mathbf{u}$ is a pixel in $I_i$, and a small residual indicates that the current flow prediction is geometrically consistent with the BA alignment.

\begin{algorithm}[t]
\caption{Reliability-aware Flow Refinement at Iter $t$.}
\label{alg:masked_flow_refine}
\begin{algorithmic}[1]
\STATE $corr = \text{4D Correlation Volume}(F^t_{i \rightarrow j}+u_i)$
\STATE $corr = corr \cdot M_i$
\STATE $inputs = \text{FeatureAggregator}(corr, inp_i, F^t_{i \rightarrow j}, M_i)$
\STATE $\Delta F^t_{i \rightarrow j} = \text{FlowGRU}(inputs)$
\STATE $F^{t+1}_{i \rightarrow j} = F^t_{i \rightarrow j} + \Delta F^t_{i \rightarrow j}$
\end{algorithmic}
\end{algorithm}

\noindent\textbf{Node-wise Geometric Reliability.}
To incorporate a more global consistency check, we compute the average geometry residual $\mathcal{L}_{\text{geo}}$ across all neighbors $\mathcal{N}(i)$ of keyframe $I_i$:
\begin{equation}
M^{\text{node}}_i(\mathbf{u}) =
\begin{cases}
1 & \text{if } \frac{1}{n} \sum_{j \in \mathcal{N}(i)} \mathcal{L}_{\text{geo}}^{(i,j)}(\mathbf{u}) < \tau_{\text{node}}, \\
0 & \text{otherwise},
\end{cases}
\end{equation}
where $n = |\mathcal{N}(i)|$. This node-wise indicator provides a global confidence estimate by assessing the geometric agreement of each pixel across views.

\begin{table*}[!t]
\centering
\setlength{\tabcolsep}{1.8mm}
\begin{tabular}{l|ccccccccc|c@{}}

& 360            & Desk           & Desk2          & Floor          & Plant          & Room           & Rpy            & Teddy          & Xyz            & Avg.           \\ \midrule
ORB-SLAM3   & -              & 0.017          & 0.210          & -              & 0.034          & -              & -             & -              & 0.009          & -              \\
DeepV2D     & 0.243          & 0.166          & 0.379          & 1.653          & 0.203          & 0.246          & 0.105          & 0.316          & 0.064          & 0.375          \\
DeepFactors & 0.159          & 0.170          & 0.253          & 0.169          & 0.305          & 0.364          & 0.043          & 0.601          & 0.035          & 0.233          \\
DPV-SLAM    & 0.112          & 0.018          & 0.029          & 0.057          & 0.021          & 0.330          & 0.030          & 0.084          & \colorbox[HTML]{D8E8C5}{0.010}          & 0.076          \\
DPV-SLAM++  & 0.132          & 0.018          & 0.029          & 0.050          & 0.022          & 0.096          & 0.032          & 0.098          & \colorbox[HTML]{D8E8C5}{0.010}          & 0.054          \\
GO-SLAM     & 0.089          & \colorbox[HTML]{D8E8C5}{0.016}          & \colorbox[HTML]{D8E8C5}{0.028}          & \colorbox[HTML]{FFF3BB}{0.025}          & 0.026          & \colorbox[HTML]{FFF3BB}{0.052}          & \colorbox[HTML]{D8E8C5}{0.019}          & 0.048          & \colorbox[HTML]{D8E8C5}{0.010}          & \colorbox[HTML]{FFF3BB}{0.035}          \\
DROID-SLAM  & 0.111          & 0.018          & 0.042          & \colorbox[HTML]{D8E8C5}{0.021}          & \colorbox[HTML]{D8E8C5}{0.016}          & \colorbox[HTML]{D8E8C5}{0.049}          & \colorbox[HTML]{FFF3BB}{0.026}          & 0.048          & 0.012          & 0.038          \\
VGGT-SLAM* & \colorbox[HTML]{FFF3BB}{0.071}          & 0.025          & 0.040          & 0.141          & 0.023          & 0.102          & 0.030          & \colorbox[HTML]{D8E8C5}{0.034}          & 0.014          & 0.053          \\
MASt3R-SLAM & \colorbox[HTML]{B7D3B7}{\textbf{0.049}} & \colorbox[HTML]{D8E8C5}{0.016}          & \colorbox[HTML]{B7D3B7}{\textbf{0.024}} & \colorbox[HTML]{FFF3BB}{0.025}          & \colorbox[HTML]{FFF3BB}{0.020}          & 0.061          & 0.027          & \colorbox[HTML]{FFF3BB}{0.041}          & \colorbox[HTML]{B7D3B7}{\textbf{0.009}}          & \colorbox[HTML]{D8E8C5}{0.030}          \\
\midrule
\textbf{FoundationSLAM (Ours)} & \colorbox[HTML]{D8E8C5}{0.055}          & \colorbox[HTML]{B7D3B7}{\textbf{0.015}} & \colorbox[HTML]{D8E8C5}{0.028}          & \colorbox[HTML]{B7D3B7}{\textbf{0.020}} & \colorbox[HTML]{B7D3B7}{\textbf{0.014}} & \colorbox[HTML]{B7D3B7}{\textbf{0.038}} & \colorbox[HTML]{B7D3B7}{\textbf{0.015}} & \colorbox[HTML]{B7D3B7}{\textbf{0.018}} & \colorbox[HTML]{B7D3B7}{\textbf{0.009}} & \colorbox[HTML]{B7D3B7}{\textbf{0.024}} \\ \bottomrule
\end{tabular}
\caption{Tracking accuracy on TUM-RGBD dataset. *means using uncalibrated images. }
\label{tab:tum_pose}
\end{table*}

\begin{table*}[t]
\centering
\setlength{\tabcolsep}{0.7mm}

\begin{tabular}{l|ccccccccccc|c}
     
& MH01           & MH02           & MH03           & MH04           & MH05           & V101           & V102           & V103           & V201           & V202           & V203           & Avg.           \\
\midrule
ORB-SLAM3   & 0.071          & 0.067          & 0.071          & 0.082          & 0.060          & 0.015          & 0.020          & -              & 0.021          & 0.018          & -              & -              \\
DeepV2D     & 0.739          & 1.144          & 0.752          & 1.492          & 1.567          & 0.981          & 0.801          & 1.570          & 0.290          & 2.202          & 2.743          & 1.298          \\
DeepFactors & 1.587          & 1.479          & 3.139          & 5.331          & 4.002          & 1.520          & 0.679          & 0.900          & 0.876          & 1.905          & 1.021          & 2.040          \\
DPV-SLAM    & \colorbox[HTML]{D8E8C5}{0.013}          & 0.016          & \colorbox[HTML]{FFF3BB}{0.022}          & \colorbox[HTML]{D8E8C5}{0.043}          & \colorbox[HTML]{D8E8C5}{0.041}          & \colorbox[HTML]{D8E8C5}{0.035}          & \colorbox[HTML]{B7D3B7}{\textbf{0.008}}          & \colorbox[HTML]{B7D3B7}{\textbf{0.015}}          & 0.020          & \colorbox[HTML]{D8E8C5}{0.011}          & 0.040          & 0.024          \\
DPV-SLAM++  & \colorbox[HTML]{D8E8C5}{0.013}          & 0.016          & \colorbox[HTML]{D8E8C5}{0.021}          & \colorbox[HTML]{B7D3B7}{\textbf{0.041}}          & \colorbox[HTML]{D8E8C5}{0.041}          & \colorbox[HTML]{D8E8C5}{0.035}          & \colorbox[HTML]{FFF3BB}{0.010}          & \colorbox[HTML]{B7D3B7}{\textbf{0.015}}          & 0.021          & \colorbox[HTML]{D8E8C5}{0.011}          & 0.023          & \colorbox[HTML]{FFF3BB}{0.023}          \\
GO-SLAM     & 0.016          & \colorbox[HTML]{D8E8C5}{0.014}          & 0.023          & 0.045          & 0.045          & 0.037          & 0.011          & 0.023          & \colorbox[HTML]{D8E8C5}{0.016}          & \colorbox[HTML]{B7D3B7}{\textbf{0.010}}          & \colorbox[HTML]{FFF3BB}{0.022}          & 0.024          \\
DROID-SLAM  & \colorbox[HTML]{D8E8C5}{0.013}          & \colorbox[HTML]{D8E8C5}{0.014}          & \colorbox[HTML]{FFF3BB}{0.022}          & \colorbox[HTML]{D8E8C5}{0.043}          & 0.043          & 0.037          & 0.012          & 0.020          & \colorbox[HTML]{FFF3BB}{0.017}          & 0.013          & \colorbox[HTML]{D8E8C5}{0.014}          & \colorbox[HTML]{D8E8C5}{0.022}          \\
MASt3R-SLAM & 0.023          & 0.017          & 0.057          & 0.113          & 0.067          & 0.040          & 0.019          & 0.027          & 0.020          & 0.025          & 0.043          & 0.041          \\ \midrule
\textbf{FoundationSLAM (Ours)}  & \colorbox[HTML]{B7D3B7}{\textbf{0.010}} & \colorbox[HTML]{B7D3B7}{\textbf{0.011}} & \colorbox[HTML]{B7D3B7}{\textbf{0.020}} & \colorbox[HTML]{B7D3B7}{\textbf{0.041}} & \colorbox[HTML]{B7D3B7}{\textbf{0.040}} & \colorbox[HTML]{B7D3B7}{\textbf{0.034}} & \colorbox[HTML]{D8E8C5}{0.009} & \colorbox[HTML]{B7D3B7}{\textbf{0.015}} & \colorbox[HTML]{B7D3B7}{\textbf{0.014}} & \colorbox[HTML]{B7D3B7}{\textbf{0.010}} & \colorbox[HTML]{B7D3B7}{\textbf{0.011}} & \colorbox[HTML]{B7D3B7}{\textbf{0.019}} \\ \bottomrule
\end{tabular}
\caption{Tracking accuracy on EuRoC dataset.}
\label{tab:euroc_pose}
\end{table*}

\noindent\textbf{Reliability Adaptive Refinement.}  
We combine the local and global indicators into a unified binary reliability mask:
\begin{equation}
M_i(\mathbf{u}) = M^{\text{edge}}_{i \rightarrow j}(\mathbf{u}) \cdot M^{\text{node}}_i(\mathbf{u}).
\end{equation}

During flow refinement, this mask defines two distinct update strategies:

1. \textit{Reliable regions} ($M_i = 1$): Flow updates rely on local correlation volumes, assuming the match lies within a narrow search window. This enables efficient refinement using high-resolution correlation sampling.

2. \textit{Unreliable regions} ($M_i = 0$): We mask out correlation features and remove them from the update pipeline. Flow updates are instead driven by contextual information containing geometry priors, enabling robust correction in ambiguous regions.

Prior works often incorporate optimization residuals into flow refinement by simply feeding them into the predictor as additional input feature~\cite{teed2021droid}. Despite some optimization feedback, it does not alter the flow estimation process structurally. All regions, reliable or not, still rely on local correlation-based matching, which can lead to persistent errors in ambiguous regions.

In contrast, our approach explicitly separates the refinement behaviors through a structured, reliability-aware design. By masking out correlation features in unreliable regions, we force the network to rely solely on geometry-guided context for refinement, thus learning to handle difficult areas more effectively. For reliable regions, correlation features are retained, ensuring both precision and efficiency. This selective refinement improves learning dynamics, robustness, and overall SLAM performance.

\noindent\textbf{Implementation Notes.}
In practice, we set both $\tau_{\text{edge}} = 5$ and $\tau_{\text{node}} = 5$, which are slightly larger than the correlation search radius of 3. This ensures that residuals outside the scope of reliable cost volume matching are excluded, aligning with the intended division of refinement responsibilities across reliable and unreliable regions.

\section{Experimental Results}

In this section, we evaluate FoundationSLAM on multiple public benchmarks, demonstrating its effectiveness in both localization and dense reconstruction.

\noindent\textbf{Implementation Details.}  
We train our model on 6-frame sequences sampled from the TartanAir~\cite{wang2020tartanair} dataset, following ~\cite{teed2021droid}. Each sequence forms a co-visibility graph with 18 edges. This sampling strategy is only used during training and does not affect the online keyframe selection during inference. We resize input images to $512 \times 384$. The model is trained for 300K steps using AdamW optimizer with OneCycleLR scheduling, a learning rate of $3.5 \times 10^{-4}$, and weight decay of $10^{-5}$. Training takes approximately 5 days on 8 RTX 4090 GPUs with a batch size of 8. At test time, with an efficient design adopting ViT-S and running foundation encoding on half-resolution, the system runs at 18 FPS.

\noindent\textbf{Baselines}. We compare with state-of-the-art SLAM systems representing key paradigms: classical sparse methods (ORB-SLAM3~\cite{campos2021orb}), matching-based approaches (DeepV2D~\cite{teed2018deepv2d}, DeepFactors~\cite{czarnowski2020deepfactors}, DPV-SLAM~\cite{lipson2024deep}, DROID-SLAM~\cite{teed2021droid}), the NeRF based approach GO-SLAM~\cite{zhang2023go}, and recent geometry-enhanced techniques VGGT-SLAM~\cite{wang2025vggt}, MASt3R-SLAM~\cite{Murai_2025_CVPR}.

\begin{figure*}[!t]
\centering
\includegraphics[width=\linewidth]{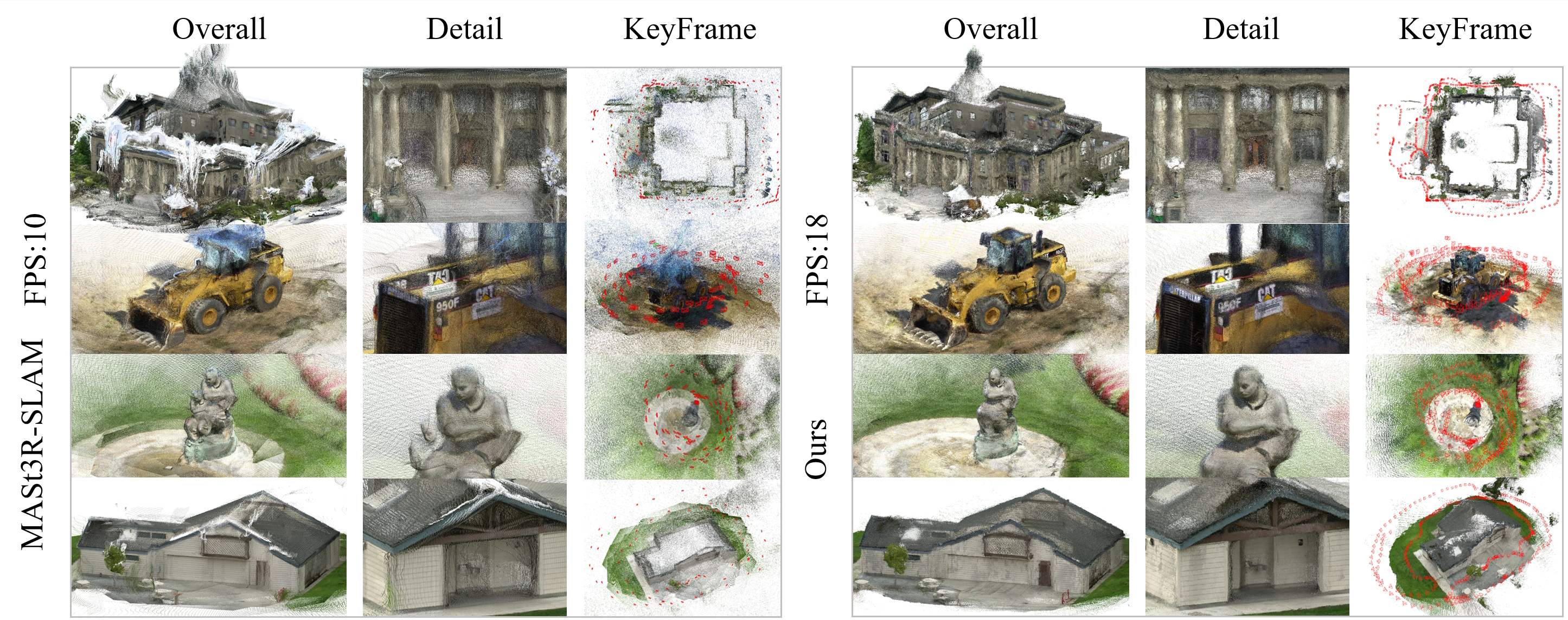}
\caption{Qualitative Comparison on TNT Dataset. We show the qualitative results of our method and MASt3R-SLAM on the TNT~\cite{tnt} dataset, including the overall and detailed reconstruction, keyframe trajectories. Our method maintains significantly more keyframes while ensuring better geometric consistency, with less layering and artifacts.}
\label{fig:demo}
\end{figure*}

\subsection{Tracking Evaluation}

We evaluate our system on three standard benchmarks: TUM-RGBD, EuRoC MAV, and ETH3D-SLAM. As shown in Tables~\ref{tab:tum_pose} and~\ref{tab:euroc_pose}, FoundationSLAM achieves state-of-the-art ATE RMSE across all datasets. On TUM-RGBD, it ranks first on 7 out of 9 sequences, with particularly strong performance in reflective or low-texture environments. On EuRoC, which involves grayscale drone footage with rapid motion and significant domain shift, our method outperforms both DROID-SLAM and MASt3R-SLAM, demonstrating high robustness to viewpoint variation. ETH3D-SLAM further showcases its stability under severe motion blur and dynamic scenes. As reported in Table~\ref{tab:tracking-eth3d}, our method achieves the highest AUC across error thresholds, indicating consistent localization quality over varying conditions.

\begin{table}[t]
\centering
\setlength{\tabcolsep}{11pt}

\begin{tabular}{l|cc}

\toprule
Method & ATE & AUC   \\ \midrule
ORB-SLAM3  & 0.135 & 16.661 \\
DROID-SLAM  & 0.171 & 22.297 \\
DPVO  & 0.137 & 22.628 \\
DPV-SLAM  & \colorbox[HTML]{FFF3BB}{0.109} & \colorbox[HTML]{FFF3BB}{23.097} \\
DPV-SLAM++  & 0.132 & 21.784 \\
MASt3R-SLAM & \colorbox[HTML]{D8E8C5}{0.086} & \colorbox[HTML]{D8E8C5}{23.935} \\

\midrule
\textbf{FoundationSLAM (Ours)}    & \colorbox[HTML]{B7D3B7}{\textbf{0.069}} & \colorbox[HTML]{B7D3B7}{\textbf{24.775}} \\ \bottomrule
\end{tabular}

\caption{Tracking accuracy on ETH3D-SLAM dataset.}
\label{tab:tracking-eth3d}
\end{table}

\begin{figure}[!t]
\centering
\includegraphics[width=\linewidth]{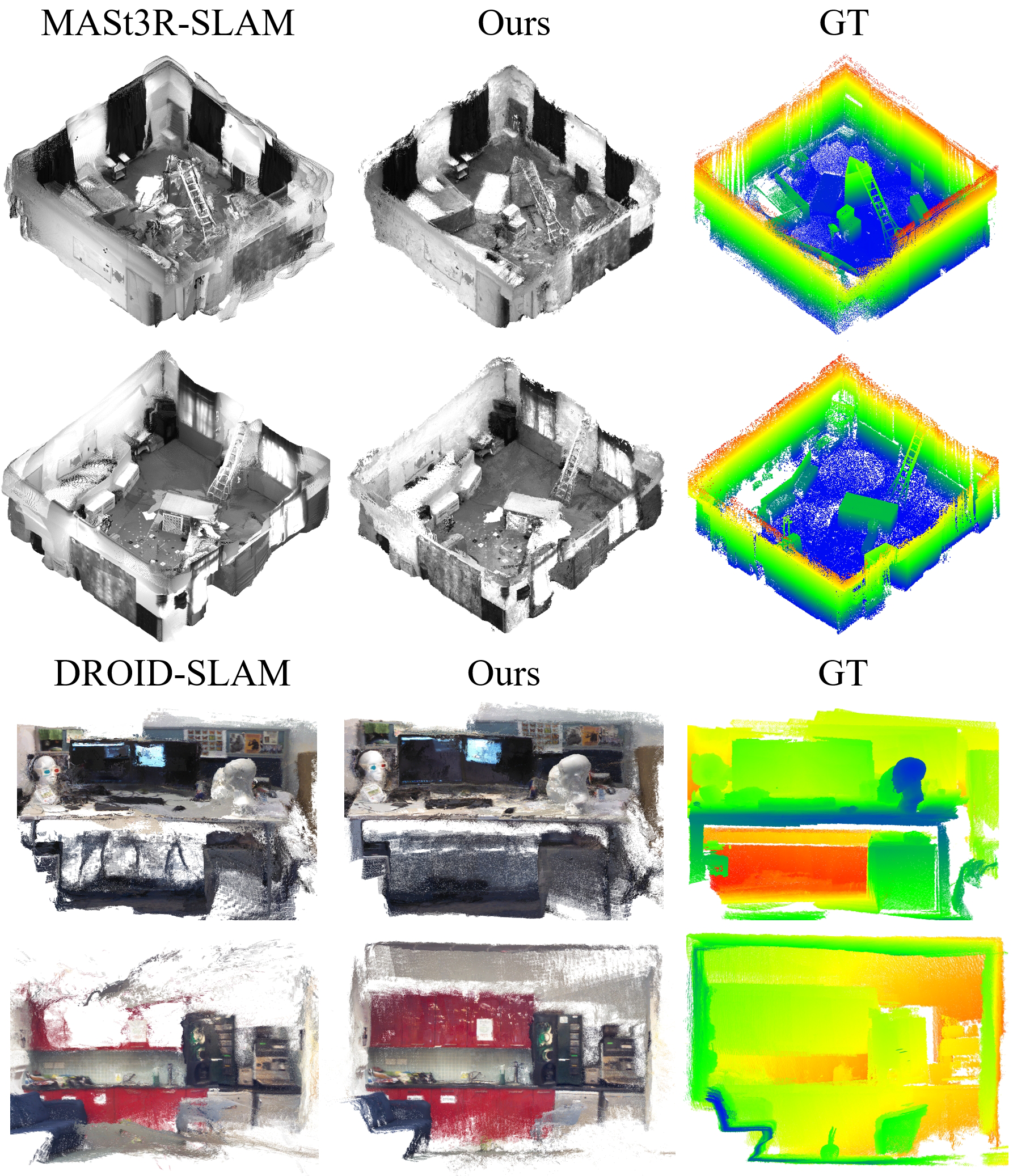}
\caption{Qualitative Reconstruction Comparison. Comparison with SOTA baselines on EuRoC and 7Scenes.}
\label{fig:qualitative}
\end{figure}
\subsection{Mapping Evaluation}

We evaluate the dense reconstruction performance on 7Scenes (seq-01) and EuRoC (VICON room sequences), comparing against DROID-SLAM, MASt3R-SLAM, and VGGT-SLAM. Evaluation metrics include \textit{Accuracy} (average distance from each reconstructed point to its closest ground-truth point), \textit{Completion} (average distance from each ground-truth point to its nearest reconstructed point), and \textit{Chamfer Distance} (the symmetric average of the two). For all metrics, distances are clipped at a maximum threshold of 0.5 meters to avoid the influence of large outliers.

\begin{table}[!t]
\centering
\setlength{\tabcolsep}{1.6pt}

\begin{tabular}{l|cccc}

Method    & ATE   & Acc.  & Comp.  & Cf. \\ \midrule
DROID-SLAM  & \colorbox[HTML]{FFF3BB}{0.049} & \colorbox[HTML]{D8E8C5}{0.052}    & 0.076       & \colorbox[HTML]{FFF3BB}{0.064}   \\
MASt3R-SLAM & \colorbox[HTML]{D8E8C5}{0.047} & 0.074    & \colorbox[HTML]{D8E8C5}{0.057}       & 0.066   \\
VGGT-SLAM*   & 0.067 & \colorbox[HTML]{D8E8C5}{0.052}    & \colorbox[HTML]{FFF3BB}{0.058}       & \colorbox[HTML]{D8E8C5}{0.055}   \\
\textbf{FoundationSLAM (Ours)} & \colorbox[HTML]{B7D3B7}{\textbf{0.043}} & \colorbox[HTML]{B7D3B7}{\textbf{0.039}}    & \colorbox[HTML]{B7D3B7}{\textbf{0.055}}       & \colorbox[HTML]{B7D3B7}{\textbf{0.047}}   \\ \bottomrule
\multicolumn{5}{c}{} \\

Method       & ATE   & Acc.  & Comp.  & Cf. \\ \midrule
DROID-SLAM  & \colorbox[HTML]{D8E8C5}{0.022} & \colorbox[HTML]{D8E8C5}{0.059}    & \colorbox[HTML]{D8E8C5}{0.070}        & \colorbox[HTML]{D8E8C5}{0.065}   \\
MASt3R-SLAM & \colorbox[HTML]{FFF3BB}{0.041} & \colorbox[HTML]{FFF3BB}{0.099}    & \colorbox[HTML]{FFF3BB}{0.071}       & \colorbox[HTML]{FFF3BB}{0.085}   \\
\textbf{FoundationSLAM (Ours)} & \colorbox[HTML]{B7D3B7}{\textbf{0.019}}  & \colorbox[HTML]{B7D3B7}{\textbf{0.035}}    & \colorbox[HTML]{B7D3B7}{\textbf{0.063}}       & \colorbox[HTML]{B7D3B7}{\textbf{0.048}}   \\ \bottomrule
\end{tabular}

\caption{Tracking and mapping accuracy on 7Scenes (top) and EuRoC (bottom) datasets. *means using uncalibrated images.}
\label{tab:mapping}
\end{table}

\begin{table}[t]
\centering
\setlength{\tabcolsep}{5pt}
\begin{tabular}{@{}cccc@{}}
\toprule
DROID-SLAM & MASt3R-SLAM & VGGT-SLAM & Ours \\ \midrule
24         & 10          & 26        & 18   \\ \bottomrule
\end{tabular}
\caption{Comparison of frames per second on EuRoC.}
\label{tab:fps}
\end{table}

As shown in Table~\ref{tab:mapping}, our method achieves the best overall reconstruction performance on both datasets. On the EuRoC VICON room sequences, our method surpasses DROID-SLAM in both accuracy and completion, resulting in a lower Chamfer distance. In contrast, MASt3R-SLAM shows poor reconstruction quality, likely due to the domain gap caused by its lack of training on grayscale data. These results demonstrate the robustness of our method in fast-motion, grayscale scenes with wide baselines.
On the 7Scenes dataset, our approach improves reconstruction accuracy over VGGT-SLAM and MASt3R-SLAM. While all methods yield similar completion scores, our method still achieves the lowest Chamfer distance, indicating better overall geometry alignment. These results further validate the potential of the optical flow-based SLAM paradigm with end-to-end bundle adjustment for dense reconstruction, even when compared against methods that incorporate strong geometry priors.
Figure~\ref{fig:demo} provides a direct comparison with MASt3R-SLAM on the TNT dataset, demonstrating our method's reconstruction advantages. Additional qualitative comparisons are shown in Figure~\ref{fig:qualitative}, highlighting sharper geometry and fewer outliers than competing methods.

\noindent\textbf{Inference Speed.} We evaluate the inference speed on EuRoC dataset in Table \ref{tab:fps} on a single 4090 GPU. MASt3R-SLAM maintains significantly fewer keyframes, as shown in Figure~\ref{fig:demo}, while VGGT-SLAM outputs tracking results per submap rather than per frame, contributing to their speed advantages. Our system achieves real-time inference at 18 FPS, striking a balance between performance and efficiency.

\begin{table}[t]
\centering
\setlength{\tabcolsep}{5.0pt}
\begin{tabular}{ccc|cccc}
\toprule
Bi-BA & $M_{\text{node}}$ & $M_{\text{edge}}$ & ATE   & Acc.  & Comp. & Cf.   \\ \midrule
   &    &    & 0.021 & 0.051 & 0.070 & 0.061 \\ \midrule
   & \checkmark &  & 0.021 & 0.047 & 0.067 & 0.057 \\
   &  & \checkmark & 0.021 & 0.048 & 0.068 & 0.058 \\ 
   & \checkmark & \checkmark & 0.020 & 0.046 & 0.066 & 0.056 \\ \midrule
\checkmark & \checkmark &    & 0.020 & 0.036 & 0.065 & 0.051 \\
\checkmark &    & \checkmark & 0.020 & 0.038 & 0.067 & 0.052 \\ \midrule
\checkmark & \checkmark & \checkmark & \textbf{0.019} & \textbf{0.035} & \textbf{0.063} & \textbf{0.048} \\ \bottomrule
\end{tabular}
\caption{Ablation Studies on EuRoC datasets. Results demonstrate the contribution of each proposed component to localization and reconstruction performance in SLAM.}
\label{tab:ablation}
\end{table}

\subsection{Ablation Studies}
In Table~\ref{tab:ablation}, we conduct ablation studies on the EuRoC dataset to evaluate the impact of our Bi-Consistent BA Layer and Reliability-Aware Refinement components.

\begin{figure}[!t]
\centering
\includegraphics[width=\linewidth]{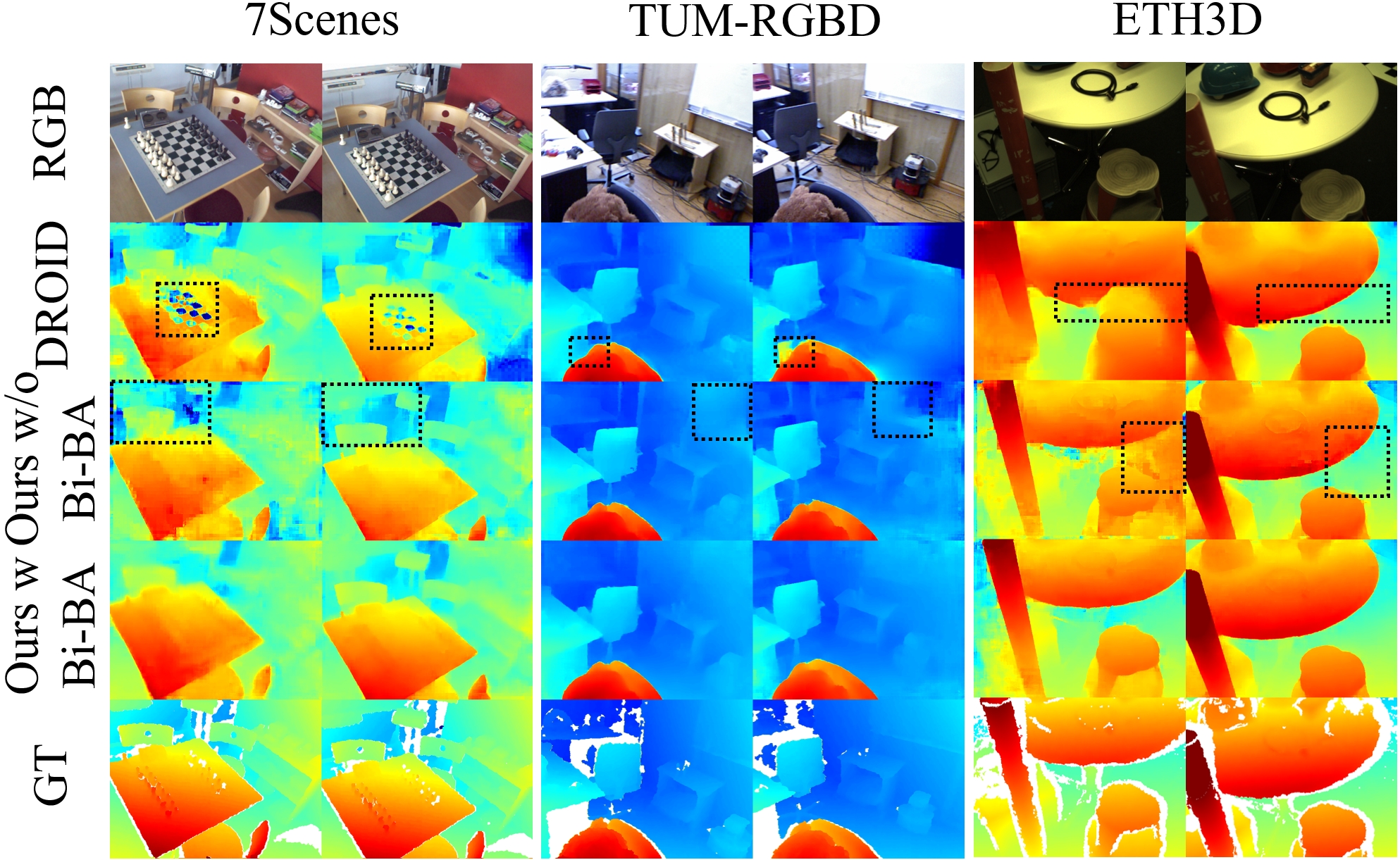}
\caption{Impact of Bi-Consistent BA Layer. We visualize the depth maps of neighboring keyframes to reveal inconsistencies caused by unreliable optical flow in baseline systems. Our Bi-Consistent BA Layer significantly improves inter-frame geometric alignment.}
\label{fig:Bi-BA ablation}
\end{figure}

\noindent\textbf{Bi-Consistent BA Layer.}  
As shown in Figure~\ref{fig:Bi-BA ablation}, incorporating the Bi-Consistent Bundle Adjustment Layer leads to significantly improved geometric consistency across keyframes. This consistency helps produce more coherent depth and pose estimates, which in turn improves both localization and mapping performance. The quantitative improvements in Table~\ref{tab:ablation} confirm that enforcing multi-view geometric alignment is critical for dense SLAM.

\begin{figure}[!t]
\centering
\includegraphics[width=\linewidth]{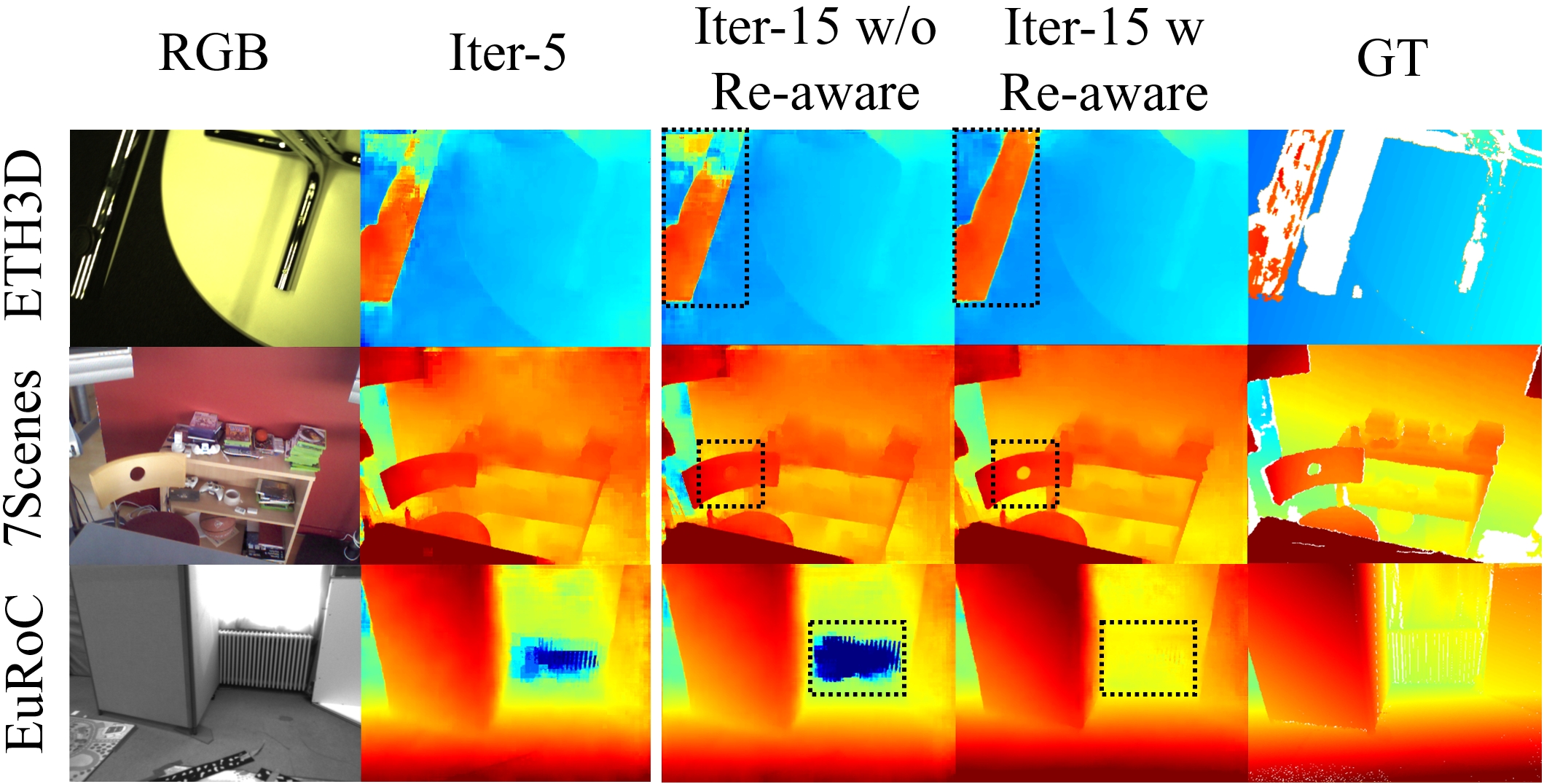}
\caption{Flow Refinement on Challenging Areas. We visualize the flow refinement process in challenging regions across three datasets. Effect of the Reliability-Aware Refinement mechanism for correcting difficult-to-match areas such as holes, reflections, and repeated textures.}
\label{fig:Re-aware-ablation}
\end{figure}

\noindent\textbf{Reliability-Aware Flow Refinement.}  
Our refinement module introduces two reliability modeling strategies: Node-wise Geometric reliability and Edge-wise Flow reliability. Both contribute individually, and their combination yields further gains, as reflected in Table~\ref{tab:ablation}. Figure~\ref{fig:Re-aware-ablation} provides qualitative examples, where ours corrects errors in reflective, low-texture, or ambiguous regions that challenge correlation-based methods. These results highlight the importance of reasoning about flow reliability under difficult conditions.

\noindent\textbf{Full Model.}  
Combining Bi-consistent BA with both reliability modules yields the best overall performance. This demonstrates the benefit of tight integration between flow estimation and geometry reasoning, and highlights the importance of enforcing consistency and reliability throughout the SLAM optimization loop.

\section{Conclusions}

We present FoundationSLAM, an end-to-end monocular dense SLAM system that integrates geometry-guided optical flow estimation with multi-view consistent optimization. Our framework unifies three key components: a Hybrid Flow Network enhanced by foundation depth priors, a Bi-Consistent Bundle Adjustment Layer enforcing cross-view consistency, and a Reliability-Aware Refinement mechanism guided by geometric residuals. By tightly integrating optical flow estimation with geometry-aware optimization, FoundationSLAM achieves state-of-the-art performance in tracking and mapping across challenging benchmarks. Running in real time on monocular RGB input, our method demonstrates strong generalization and robustness, enabling practical use in diverse real-world scenarios.

\section*{Acknowledgements}
This work is supported by the Key R\&D Program of Jiangxi Province, China (20232BBE50019), and the National Natural Science Foundation of China (62276016, 62372029).

\bibliography{aaai2026}

\end{document}